\documentclass[conference]{IEEEtran}
\usepackage{cite}
\ifCLASSINFOpdf
  \usepackage[pdftex]{graphicx}
\else
\fi
\usepackage{amsmath}
\usepackage{url}
\usepackage{tabularx}
\usepackage{multirow}

\newcommand{\entropy}{$\delta$E}
\newcommand{\similarity}{SSIM}
\newcommand{\encerr}{EE}
\newcommand{\cornerr}{CE}

\newcommand{\TRAINMETRICWIDTH}{.9\columnwidth}

\begin{document}
\title{DOOM Level Generation using Generative Adversarial Networks}

\author{
\IEEEauthorblockN{Edoardo Giacomello}
\IEEEauthorblockA{Dipartimento di Elettronica,\\
Informazione e Bioinformatica\\
Politecnico di Milano\\
edoardo.giacomello@mail.polimi.it}
\and
\IEEEauthorblockN{Pier Luca Lanzi}
\IEEEauthorblockA{Dipartimento di Elettronica,\\
Informazione e Bioinformatica\\
Politecnico di Milano\\
%Milano, Italy.\\
pierluca.lanzi@polimi.it}
\and
\IEEEauthorblockN{Daniele Loiacono}
\IEEEauthorblockA{Dipartimento di Elettronica,\\
Informazione e Bioinformatica\\
Politecnico di Milano\\
%Milano, Italy.\\
daniele.loiacono@polimi.it}}


\maketitle

\begin{abstract}
%%% TEXEXPAND: INCLUDED FILE MARKER ./abstract.tex
We applied Generative Adversarial Networks (GANs) to learn a model of DOOM levels
from human-designed content. Initially, we analyzed the levels and
	extracted several topological features.
Then, for each level, we extracted a set of images identifying the occupied
area, the height map, the walls, and the position of game objects.
%%%
We trained two GANs: one using plain level images, one using both the images
	and some of the features extracted during the preliminary analysis.
We used the two networks to generate new levels and compared the results to
	assess whether the network trained using also the topological features
	could generate levels more similar to human-designed ones.
%%%
Our results show that GANs can capture intrinsic structure of DOOM levels
	and appears to be a promising approach to level generation in first person shooter games.
%%% TEXEXPAND: END FILE ./abstract.tex
\end{abstract}


% \fi
\IEEEpeerreviewmaketitle

\section{Introduction}
\label{sec:introduction}
%%% TEXEXPAND: INCLUDED FILE MARKER ./sec_introduction.tex
% !TEX root=main.tex
Content creation is nowadays one of the most expensive and time consuming tasks in the game development process.
Game content can be either \emph{functional} or \emph{non-functional}~\cite{pcgpaper};
functional content, such as weapons, enemies, and levels, is related to the game mechanics and directly affects game dynamics;
non-functional content, such as textures, sprites, and 3D models, is not related to game mechanics and has  a limited impact on
  game dynamics.
In this context, levels are of paramount importance, especially in first person shooter and platform games, as they greatly affect
  the player experience.
Unfortunately, level design usually heavily relies on domain expertise, good practices, and an extensive playtesting.
To deal with these issues, several game researchers are spending considerable effort on studying and designing procedural content generation
  systems that, exploiting machine learning and search algorithms, can model the level design process and assist human designer.

In this work, we focus on the level design for DOOM\footnote{\url{https://en.wikipedia.org/wiki/Doom_(franchise)}}, a first person shooter game
  released in 1993 that is considered a milestone in video game history and today still has an active community of players.
There are several collections of DOOM levels freely available online,
%Interestingly, it is also possible to find online several collections of DOOM levels,
	like the \emph{Video Game Level Corpus}~\cite{VGLC} (VGLC),
	which includes the official levels of DOOM and DOOM2 represented in multiple formats, and
	the \emph{idgames archive}\footnote{\url{http://doom.wikia.com/wiki/Idgames_archive}}, a large repository with more than 9000 DOOM
	levels created by the community.
Thanks to the possibility offered by such publicly available data,
	in this paper we propose and study a novel method for the procedural generation of DOOM levels using Generative Adversarial Networks (GANs)
	a type of deep neural network.
%%%
%We trained two different models of Generative Adversarial Networks (GANs),
%	a type of deep neural networks, on a dataset of images existing DOOM levels;
%	then, sampled the trained networks to generate new levels and evaluated the results.
%%%%
%To test our approach,
Initially, we created a dataset suitable for training a GAN from more than 1000 DOOM levels.
We processed each level to generate (i) a set of images that represent the
  most important features of the level (i.e., the walkable area, walls, floor height, objects, and room segmentation), and (ii) a vector of
  numerical and categorical attributes that effectively describe the level (e.g., size, length of the perimeter, number of rooms, etc.).
Then we trained two models of GAN: (i) an \emph{unconditional} GAN that uses as input only the images of the levels and (ii) a \emph{conditional}
  GAN that uses as input also a selection of the features extracted from the existing levels.
To evaluate and compare the quality of the levels generated by the two networks we designed a set of metrics, inspired to the ones used
  to evaluate the quality of indoor maps created by SLAM algorithms~\cite{slam}.
Our results show that the quality of the generated levels improves during the training process and, after 36000 iterations, both networks
   generate levels of good visual quality and with a limited amount of noise.
The results also suggest that the \emph{conditional} GAN is able to exploit the input feature vector to generate levels with a slightly
  higher visual and structural similarity to human-designed levels.
%%% TEXEXPAND: END FILE ./sec_introduction.tex

\section{Related Work}
\label{sec:related}
%%% TEXEXPAND: INCLUDED FILE MARKER ./sec_related.tex
% !TEX root=main.tex
\subsection{Generative Adversarial Networks}
\begin{figure}
	\begin{center}
		\includegraphics[width=1\linewidth]{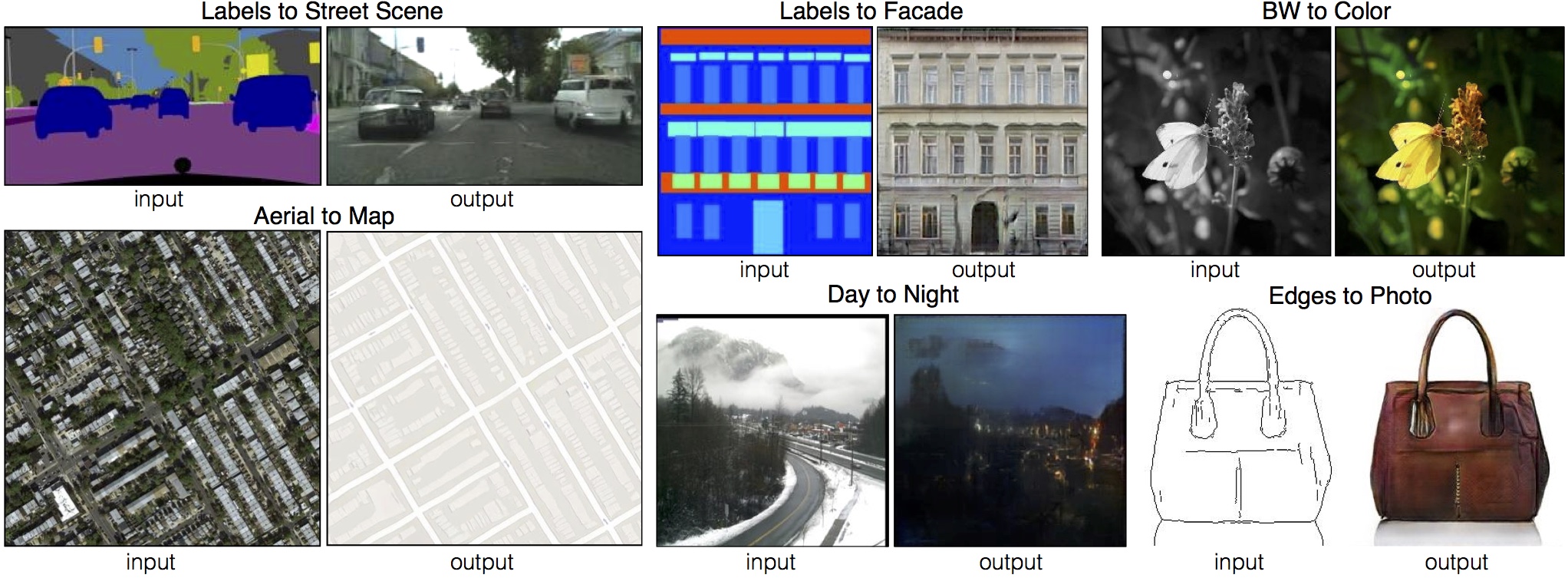}
	\end{center}
	\caption{Examples taken from \cite{image-to-image} on the image-to-image translation problem.}
	\label{fig:img-to-img}
\end{figure}

Generative Adversarial Networks (GANs) are a recent generative model based on Artificial Neural Networks.
This type of models learns the data distribution of a dataset and generate synthetic data that exhibit similar characteristics to the real data.
Among all the domains in which GANs have been already applied, image processing is one of the most prominent.
For example, Radford et al.~\cite{gan:dcgan} presented several applications of GANs that involve handwritten digits~\cite{dataset:MNIST}, human faces~\cite{dataset:celebA}, and bedrooms~\cite{dataset:LSUN}.
Then, a large amount of creative work was performed also on datasets about birds and flowers~\cite{gan:birds}.
Another successful application of GANs is the \emph{image-to-image translation}:
Isola et al.~\cite{image-to-image} investigated GANs as a general solution to this problem in several settings such as image colorization,
  segmented image to realistic scene conversion or generation of realistic objects starting from hand drawn input
	  (see for example Figure~\ref{fig:img-to-img}).
GANs have also been employed in many other domains such as frame prediction in videos~\cite{gan:frameprediction}
  and sound generation~\cite{gan:sound}.

\subsection{PCGML in Video-Games}
Procedural Content Generation (PCG) was used in early days of game development to deal with memory and computational limitations~\cite{pcgbook}.
Notable examples include \emph{Elite}~\cite{game:elite}, a space simulation in which procedural generation is used to create the game universe,
  and \emph{Rogue}~\cite{game:rogue}, a dungeon-crawling game in which dungeon rooms and the hallways are generated by means of an algorithm.
More recently, PCG is still used in several games to support the generation of content and to increase the \emph{re-playability} of games.
Some examples of PCG applications in recent games are \emph{Diablo}~\cite{game:diablo}, \emph{Minecraft}~\cite{game:minecraft},
  \emph{Dwarf Fortress}~\cite{game:dwarf}, \emph{Elite: Dangerous}~\cite{game:elitedangerous}, and \emph{No Man's Sky}~\cite{game:nomanssky}.

Recently,
%thanks to the recent development of machine learning techniques,
Summerville et al.~\cite{PCGML} introduced a novel approach to
  PCG based on machine learning (PCGML), that consists of training machine learning models (e.g., a deep neural network) on existing game
	contents and, thus, using the trained models to generate novel contents.
This approach proved to be successful in several applications~\cite{PCGML},
  %and a lot of work has been done on methods and data representation~\cite{PCGML},
  despite most of the previous works focused on the level generation for platform games.
For example, Dahlskog et al.~\cite{mariongrams} used \emph{n-grams} to generate new levels for \emph{Super Mario}\cite{game:supermario},
  Rishabh et al.~\cite{levelsautoencoder} used \emph{autoencoders}, while Snodgrass and Onta{\~n}{\'o}n~\cite{mariomarkovchains} proposed
	 an approach based on Markov Chains.
In~\cite{zeldalevels} the authors presented a method for generating levels of \emph{Legend of Zelda}\cite{game:zelda}
  using Bayes Nets for the high level topological structure and Principal Component Analysis for generating the rooms.
In contrast, in this work, we propose
	a method for generating the whole level topology using a single model,
	with the possibility of easily adding more features
	or eventually applying the same structure to another dataset.
In their work, Scott et al.\cite{resourcelocation} use neural networks for predicting resources location in \emph{StarCraft II} maps \cite{game:starcraft}.
%%%
Although the data domain is similar to the one used in our work, they focused on resource placement rather than map topology generation.
  %and requires the image of an already existing level as input.
%Moreover, we make use of GAN, which is a particular setting in which a generator is able to produce new samples using a vector of noise as input.
Beckham and Pal~\cite{heightgen} proposed a method based on GANs
	for generating realistic height level maps for video games,
	that is more focused on landscapes and might be difficult to apply to generate indoor environments.
Finally, the same approach can be also applied to non-functional content generation, as done in~\cite{spritegen}, in which GANs are used to
  generate 2d sprites.

% \subsection{Video Game Level Corpus}
% One of the major problem with generative models, as discussed in~\cite{PCGML}, is that they require a large amount of data to be optimized.
% Unfortunately, the domain of video-games levels does not benefit of large datasets to work with, and generally levels from different video games
%   does not share common data structures.
% Summerville et al.~\cite{VGLC} created the \textit{Video Game Level Corpus (VGLC)}, a collection of game levels represented in multiple formats.
% Using this format as starting point, we collected about 9000 user-generated Doom levels of different size and designed an extended representation
%  that better fit our needs, while still keeping the dataset compatible with the original VGLC representation.
% Although VGLC provides the parser they used for data generation, we wrote a new parser which better integrates with our system and feature
%  representation, and can also be used as a stand-alone parser for future researches.
%%% TEXEXPAND: END FILE ./sec_related.tex

\section{Analysis of DOOM Levels}
\label{sec:levels}
%%% TEXEXPAND: INCLUDED FILE MARKER ./sec_levels.tex
DOOM is a major milestone in the history of video games that greatly influenced the First Person Shooter game genre.
%%%
In DOOM, the player traverses a series of levels populated by enemies that she must kill to reach the exit, while collecting
	weapons, ammunition, power-ups, and keys to open doors.
%Often, in order to reach the exit point, the player has to explore the level to find some keys required to open one or more doors on the path.
%DOOM is organized in episodes of 32 levels each.
%The first episode was released as shareware and \textit{Id Software} greatly encouraged the diffusion of the game.
%This contributed to the celebrity of the game in a time in which internet was still not accessible to everyone.
%
%In this section we describe how levels of \emph{DOOM} are represented.
%First we describe the level representation used by developers, then we describe how to represent a level as
%  a set of images, and, finally as a graph.
% Finally we describe also the type of features extracted from each level to both provide additional inputs to the neural networks and perform
%   a more accurate analysis of the levels.

\subsection{Level Representation (WADs)}
\label{ssec:wad}
\begin{figure}
	\begin{center}
		\includegraphics[width=0.8\columnwidth]{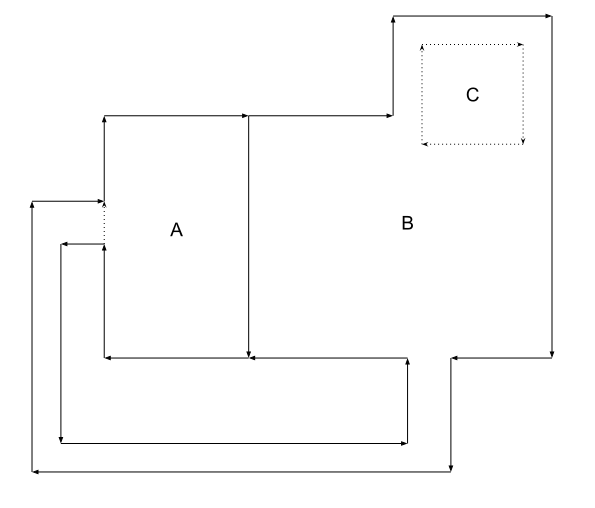}
	\end{center}
	\caption{A level with three sectors (A, B and C) and the linedefs defining them: 
		solid linedefs represent walls while dashed linedefs represent changes in height between two sectors (steps);
		thus, sector C can be viewed as a small platform inside the sector B.}
	\label{fig:sectors}
\end{figure}

%%% TEXEXPAND: INCLUDED FILE MARKER ./ssec_wad.tex
DOOM levels are saved as WAD files that store all the topological information as well as the multimedia assets. 
%
%The \emph{DOOM} Game Engine \cite{doomengine} makes use of package files called \emph{WAD} to store every game resource such as Levels, Textures, Sounds, etc.
%store all the level assets (e.g., textures, sounds) and was designed to be extensible.
%In this section we provide only a brief description of how \emph{WAD} files are structured, as an in depth description is behind the scope of this
%  work (we refer the interested reader to~\cite{doomspecs}).
%
A WAD file contains an ordered sequence of records (or \emph{lumps}).
%The most relevant ones are described in the following; we refer the interested reader to~\cite{doomspecs} for a complete list.
%%%
The \textit{Name} lump simply contains the name of the level.
The lump named \textit{Things} contains all the information about the objects included in the level that are not walls, doors, or floor tiles.
The \textit{Vertexes} lump is an ordered list of each vertex in the map.
The \textit{Linedefs} lump contains all the directed lines that connects two vertices of the map; 
	lines define walls, steps, invisible boundaries, or triggers (e.g., tripwires, switches).
The \textit{Sidedefs} lump contains the texture data for each \emph{side} of a \emph{line} and the number of \emph{sector} (i.e., an area of the level)
	enclosed by that \emph{side} of the \emph{line}.
%	the position of each \emph{side} entry in this record is used as an index by other elements of the level.
%\medskip\noindent
The lump labeled \textit{Sectors} contains the information about all the sectors that is 
	areas that have a constant height and the same texture.
%A sector is typically delimited by \emph{lines} and its data include the following information:
%tag number (used as a reference by other level elements), floor height, ceiling height, floor texture, ceiling texture,
%  light level, type of floor (e.g., whether it damages the player or not), etc.
%
%\medskip\noindent
%
Figure~\ref{fig:sectors} shows an example of a level with 3 sectors.
Note that although all the above lumps are mandatory to build a playable level, some additional lumps are added to the \emph{WAD} format
	to speed up the rendering process;  we refer the interested reader to~\cite{doomspecs} for a complete list.
%
%Such additional lumps can be automatically generated using external tools.
%In this work we used BSP (v5.2)~\cite{bsp} in the last stage of the pipeline, in order to produce playable DOOM levels.

\subsection{Image Representation}
\label{ssec:images}
%%% TEXEXPAND: INCLUDED FILE MARKER ./ssec_images.tex
To train GANs with DOOM levels,
	we generated a set of images to represent the data of WAD files.
Each image represents a specific feature of the level as a grayscale 8-bit image.
%, where each pixel encodes a square of 32x32 MU.
%
%To make \emph{DOOM} levels understendable to a neural network we represented them using a set of images,
%  dubbed as \emph{FeatureMap} images.
%Each \emph{FeatureMap} is used to represent a specific feature of the level (e.g., the walls, floor heights, things in the level, etc.) as
%  a grayscale 8-bit image, where each pixel encodes a square of 32x32 MU.
In particular, we generated six images for each level.
\textit{FloorMap} images represent the level floor encoding
	(i) the empty areas that can be traversed by the players with a white pixel and
	(ii) all the areas that cannot be traversed (e.g., obstacles, areas outside the walls)
		with a black pixel.
\textit{WallMap} images represent the level walls as white one-pixel-wide lines.
\textit{HeightMap} images represent the height of the floor in each level area;
  zero values encode the areas that cannot be traversed, other values encode all the other heights.
\textit{ThingsMap} images represent data defined by the \emph{Things} lump described before;
	things are encoded as single pixels with a specific value for each type of element
	that can be found in a level.
\textit{TriggerMap} images represent all the level triggers and encode both
	(i) the type of trigger and
	(ii) a tag identifying either the trigger or the triggered object
	(local doors, remote doors, lifts, switches and teleports).
%We identified 5 type of triggers: local doors, remote doors, lifts, switches and teleports.
%Table~\ref{tab:triggermap} shows how triggers are encoded.
%The value 0 is used to encode an empty position.
%
%\begin{table}
%\caption{Encoding used for \emph{TriggerMap}.}
%\begin{tabularx}{\columnwidth}{| c | c | X | }
%	\hline
%	\textbf{Value} & \textbf{Type} & \textbf{Description} \\
%	\hline
%	0 &	- &	Empty position. \\
%	10 &	local door & Local door that require blue key.\\
%	12 &	local door & Local door that require red key. \\
%	14 &	local door & Local door that require yellow key. \\
%	16 &	local door & Local door that does not require keys. \\
%	32+i &	remote door &	Remote door with tag $i$. \\
%	64+i &	lift &	Lift with tag $i$. \\
%	128+i &	switch & Switch that activates object with tag $i$. \\
%	192+i &	teleport &	Teleport to \emph{sector} with tag $i$. \\
%	255 &	exit &	Level exit.\\
%	\hline
%\end{tabularx}
%
%\label{tab:triggermap}
%\end{table}
%
%\medskip\noindent
\textit{RoomMap}
	images represent a room segmentation of the level computed with
	an approach similar to the one used for the analysis of indoor environments \cite{7487234}.
%an euclidean distance transform~\cite{edt} is first applied to the \emph{FloorMap}, then the local maxima are found and used to identify the room
%  center coordinates, finally, such room centers are used as markers for a Watershed algorithm~\cite{watershed} (with the negative of euclidean
%  distance computed before as basin).
%The result is a resonalbe room segmentation of the level.
%%% TEXEXPAND: END FILE ./ssec_images.tex

\subsection{Features}
\label{ssec:features}
%%% TEXEXPAND: INCLUDED FILE MARKER ./ssec_features.tex
We analyzed the levels using both the WAD files and the images generated in the previous step.
%
%	and a graph computed from the \textit{RoomMap} image
For each level, we extracted 176 features (numerical and categorical)
	divided in four major groups.
\textit{Metadata} features are based on the metadata available
	for each level, such as title, authors, description, rating, number of downloads, etc.
WAD features were extracted directly from the level WAD file
	and included the number of lines, things, sectors, vertices, size of the level, etc.
\textit{Image} features were extracted from the image representation of the level
	like, for instance, the \emph{equivalent diameter} of the level, 
	fraction of area that can be traversed, perimeter of the level,
	vertical and horizontal size of the level, etc.
\textit{Graph} features are computed from a graph representation
	extracted from the \textit{RoomMap} image by applying methods
	used to analyze indoor environments~\cite{Amigoni}; they include
	number of nodes, closeness centrality, betweenness centrality, assortativity, etc.
%%% TEXEXPAND: END FILE ./ssec_features.tex

% tolta dato che se non ricordo male non la usiamo se non per le features ma non serve descriverla come rappresentazione.
%\subsection{Graph Representation}
%\label{ssec:graph}
%\input{ssec_graph}
%%% TEXEXPAND: END FILE ./sec_levels.tex

\section{Deep Level Generation}
\label{sec:framework}
%%% TEXEXPAND: INCLUDED FILE MARKER ./sec_framework.tex
% !TEX root=main.tex
\begin{figure}[t]
	\begin{center}
		\includegraphics[width=\columnwidth]{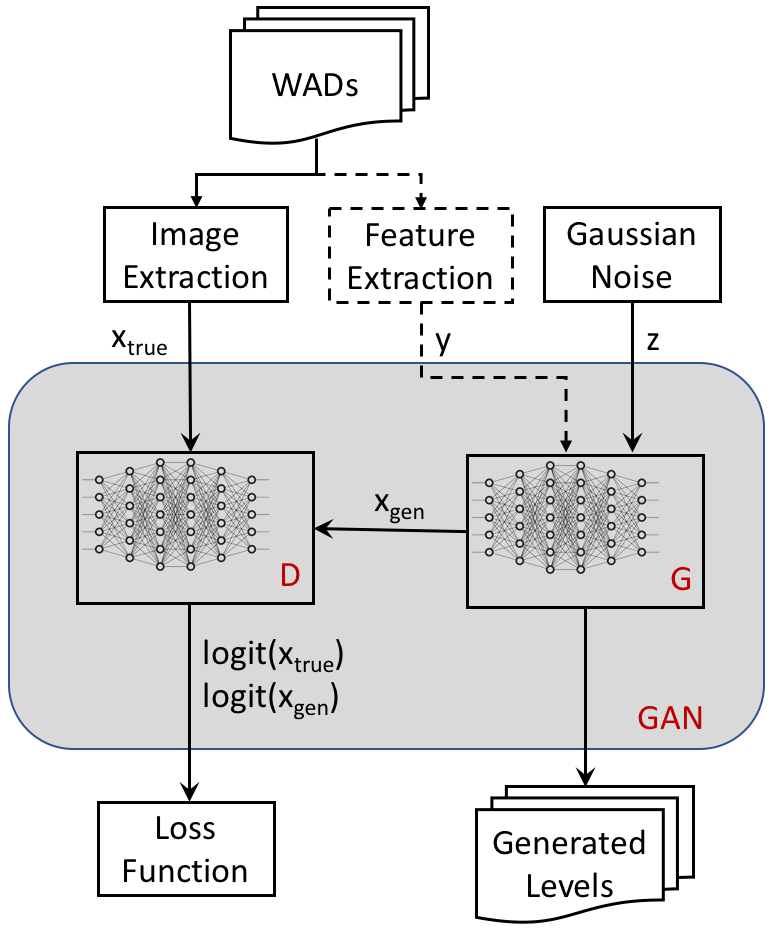}
	\end{center}
	\caption{Architecture of our framework. The GAN model (identified by the gray background) consists of two networks:
    (i) the generator network (G) and (ii) the discriminator (or critic) network (D).}
%    The feature extraction module and the input $Y$ (dashed line) are used only in the \emph{conditional} GAN model.}
	\label{fig:genmodelstructure}
\end{figure}

In this study, we applied Generative Adversarial Networks~\cite{gan} (GANs) to learn a model of existing DOOM levels;
	we then used the model to generate new levels.
Figure~\ref{fig:genmodelstructure} shows an overview of our framework. % to generate DOOM levels.
The generative network comprises two deep neural networks: a \emph{generator} network G and a \emph{discriminator} network D.
For each level, the generator network receives in input ($X_{true}$)
	(i) the six images (Section~\ref{ssec:images}) extracted from the level WAD file
	(ii) a vector of level features $Y$ (Section~\ref{ssec:features}), and
	(iii) a noise vector $Z$ sampled from a Gaussian distribution.
As a result, the GAN generates six images that represent a generated level and can be used to produce a fully playable level.
Thus, the \emph{generator} network is trained to generate an output $X_{gen}$ that is as similar as possible to the original inputs $X_{true}$.
%%%
In contrast, the discriminator network receives as input either $X_{true}$ or $X_{gen}$ (the images of
	either a level designed by a human or a level produces by the generator network).
The discriminator network outputs the \emph{logit} value of the probability that $X_{true}$ and $X_{gen}$ are images of a generated level.
Thus, the discriminator network is trained to distinguish between human-designed and generated levels.

\medskip\noindent
\textbf{Network Architecture.}
We employed the Wasserstein GAN with Gradient Penalty (WGAN-GP)~\cite{wgangp} architecture
	that proved to be more stable and to provide similar or better results than other GAN architectures~\cite{wgangp}.
%%%
In addition, we replaced the \textit{tanh} activation function on the output layer \cite{wgangp},
  with a \textit{sigmoid} function (more suitable for grayscale images with few levels of gray).
%%%
In particular, we considered two models of WGAN-GP:
	(i) an \emph{unconditional} WGAN-GP that receives as input only the images extracted by the WAD file
		and the noise vector (i.e., only $X_{true}$ and $Z$);
	(ii) a \emph{conditional} WGAN-GP that also receives the vector of features $Y$ extracted from the level.
%
% and (ii) a \emph{conditional} WGAN-GP.
%In the former model, the model receives as input only the images extracted by the WAD file of the levels and the noise vector, i.e., only $X_{true}$
%  and $Z$.
%In the latter one, the model receives as input also a vector of features extracted from the level, i.e., the inputs are $X_{true}$, $Y$, and $Z$.
%
% Among the various \emph{GAN} implementations that are proposed in the literature, we selected the Wasserstein GAN with Gradient Penalty \cite{wgangp} (WGAN-GP) described in section~\ref{sec:wgangp} as it showed better training stability with the DCGAN layer configuration and at least comparable sample quality as opposed to the other models. We considered both the unconditional and conditional versions, by parametrizing the system upon the selected input features. The only difference we introduced to the proposed model is the adoption of the \textit{sigmoid} activation function on the generator output layer in place of \textit{tanh}. This choice is motivated by the fact that tanh have been originally selected for obtaining a better colour coverage in generated rgb images, while we are interested in discrete values that often correspond to the lowest and highest output values. This showed to help the network learning faster the representation of \emph{floormap} and \emph{wallmap}, while not affecting the other maps.

\medskip\noindent
\textbf{Loss Functions.}
% We implement the Critic and Generator losses as in \cite{wgangp-imple}, which combines formulas \ref{eq:wganloss} and \ref{eq:gp}. Referencing to the notation introduced in \ref{sec:modelstructure} the losses are defined as follow:
The loss functions $L_G$ and $L_D$ used for the \emph{generator} (G) and \emph{discriminator} (D) networks are defined as,
\begin{eqnarray}
\label{eq:loss}
%\begin{split}
L_G &=& -logit(X_{Gen}) \nonumber\\
L_D &=& \underbrace{\operatorname{E}(logit(X_{Gen})) - \operatorname{E}(logit(X_{True}))}_{\text{WGAN Loss}} +
\underbrace {\lambda G_p}_{\text{Gradient Penalty}}\nonumber
%\tag{WGAN-GP Generator Loss}
%\end{split}
\end{eqnarray}
where $L_D$ and $L_G$ are the loss functions used respectively for the discriminator and the generator networks;
$X_{True}$ and $X_{Gen}$ are the images that represent human-designed and generated levels;
the gradient penalty is computed as $G_p = (\| \nabla_{\hat{X}}logit(\hat{X}) \|_2 - 1 )^2$ and $\hat{X} = \epsilon X_{True} + (1-\epsilon) X_{Gen}$,
  $\lambda = 10$, $\epsilon \sim U[0,1]$.

 \medskip\noindent
 \textbf{Training.}
We trained the discriminator and generator networks following the approach proposed in~\cite{wgangp}:
we used \emph{Adam}\cite{adam} optimizer and optimized the discriminator network five times for each update of the generator
  network.
%%%
In each training iteration, we also applied a 90 degree clockwise rotation to the input images,
	so that the networks were trained using all the four possible level orientations.
%This is sone to exploit the rotation invariance of the image representations of the level.
This transformation allows us to exploit the rotation invariance in the representation of a DOOM level,
	since its playability is not affected by its orientation in the space.

\section{Experimental Design}
\label{sec:expdesign}
%%% TEXEXPAND: INCLUDED FILE MARKER ./sec_expdesign.tex
In this paper we trained and compared two different GANs:
the first one, dubbed \emph{unconditional}, does not have any level feature in input;
the second one, dubbed \emph{conditional}, has in input a vector of features selected among the ones extracted from the level
  (Section~\ref{ssec:features}).
Both models have the same structure: the generator and the discriminator have 4 layers with 1024, 512,
  256 and 128 filters, and have been trained for 36000 iterations on the same dataset with the same learning hyperparameters 
  ($\alpha=0.0002$, $\beta_1=0$, $\beta_2=10$, and $\lambda=10$).
In the following, we briefly describe the dataset used to train the networks, how we selected the additional input features for
  the \emph{conditional} model, and how we evaluated the generated levels.

\subsection{Dataset}
The training data consists of a selection of the levels available in the
  \emph{idgames archive}, the largest online archive of levels,
  modifications, tools, and resources for the games based on the DOOM engine.
In particular, due to computational constraints, we selected only levels on one single floor and that could be represented
  with images of size not greater than 128x128 pixels.
As a result, our dataset contains 1088 levels.
Please notice that this archive and its content is not officially maintained by \emph{id Software} but it is community service.
% For preparing our experiments we filtered the DoomDataset by taking only the samples up to 128x128 in size and which had exactly one "floor".
% This led to a dataset of 1088 samples, which are then augmented by rotation during the training process.
% This is motivated from the fact that even if the level orientation does not affect playability, using levels with more floors could lead the
% network to learn a correlation between floors (and how to arrange them inside the map) which could potentially be misleading or be just enforced
% by the sample size or the way the editor arranged them on the level coordinate space. Moreover, using only one-floor levels helped in reducing
% artefacts that appeared as very small floors in resulting output.

\subsection{Feature Selection}
Because of the memory limitation of our GPUs (6Gb), we were unable to use all the 176 features we extracted for each level 
	(Section~\ref{ssec:features}) for training. 
Accordingly, we analyzed all the features and selected a small subset of them to be used as input to
	the \emph{conditional} networks.
%  
%For each level, we extracted 176 features 
%	that we analyzed to select a small subset of interesting features to provide
%In Section~\ref{ssec:features} we described the type of features extracted for each level.
%However, to keep limited the computational effort and the training complexity, we selected a small subset of features to provide as input to
%  the \emph{conditional} networks.
Features selection was guided by two principles:
(i) visual relevance, i.e., we needed features that had similar values in visually similar levels, and
(ii) robustness, i.e., we needed features not affected by noisy pixels that might be generated by the network random sampling.
At the end, we selected the following seven features:
\begin{enumerate}
	\item \emph{Equivalent diameter}: diameter of the smallest circle that encloses the whole level.
	\item \emph{Major axis length}: length of the longest axis of the level (either horizontal or vertical).
	\item \emph{Minor axis length}: length of the longest axis of the level (either horizontal or vertical).
	\item \emph{Solidity}: walkable area of the level (as a fraction of the area of the convex hull that encloses the whole level).
	\item \emph{Nodes}: number of rooms in the level.
	\item \emph{Wall Distance Skewness}: skewness of the distribution of each floor pixel distance from the closest wall; the value of this feature
    accounts for the balance between large and small areas.
	\item \emph{Wall Distance Skewness}: kurtosis of the distribution of each floor pixel distance from the closest wall; the value of this feature
    is a measure of the variety of level areas in terms of size.
\end{enumerate}

 % \begin{figure}
 % 	\includegraphics[width=\columnwidth]{feature_selection.png}
 % 	\caption{Example of feature values on a set of 5 different levels.
	% The first row shows the Room Map of the levels, in which each room is enumerated with a different grayscale colour.
	% The second row shows the feature values for the features \textit{level equivalent diameter}, \textit{level major axis length},
	% \textit{level minor axis length}, \textit{level solidity}, \textit{nodes (number of rooms)}, \textit{distmap skewness},
	% and \textit{distmap kurtosis}. }
	%  \label{fig:feature_selection}
	% \end{figure}

\subsection{Levels Evaluation}
Evaluating the quality of the samples generated from a neural network is currently an open issue~\cite{improved_gan}.
To deal with this problem, previous works~\cite{improved_gan} used either an assessment based on human annotations or on the score
  provided by the \emph{inception module}~\cite{DBLP:journals/corr/SzegedyVISW15}, an image classifier.
Unfortunately, these approaches require a lot of human effort and the \emph{inception module} cannot be easily applied to our dataset which
  is very different from ImageNet\footnote{\url{http://www.image-net.org/}} dataset it was trained on.

Accordingly, we designed a set of metrics related to the visual quality of the generated maps.
In particular, our metrics are inspired to the ones designed in~\cite{slam} to evaluate the maps generated by a SLAM algorithm running
  on a mobile agent, that are rather similar to the maps generated by our network.
Note that, the proposed metrics are not meant to provide a general solution to the problem of evaluating samples of a \emph{GAN}
  nor to improve previous work on the evaluation of maps generated by SLAM algorithms~\cite{slam}. 

\medskip\noindent
\textbf{\entropy{}}:
we compute the entropy of the pixel distribution for all the images that represent both human-designed and generated levels.
Thus, for each level image, we compute the average absolute difference between the entropy values computed for human-designed levels
  ($X_{true}$) and the ones computed for generated levels ($X_{gen}$).
As a result, this metric is able to detect whether the quantity of information encoded by the images of generated levels differs from the one
  encoded by the images of human-designed levels, as typically happens when generated images are very noisy.
  %
% This metric is computed as the average absolute difference between the entropy of \emph{FeatureMap} images
%
% 	This metric is defined as the Mean Absolute error between the entropy of two images in their colour space. In particular, since as described in chapter~\ref{sec:DatasetOrganization} we are representing maps as grey-scale images whose colour ranges between 0 and 255, we calculate the pixel distribution over each possible colour value $c$ of an image $x$, $P_{c}(x)$, then we calculate the entropy as:
% 	\begin{equation}
% 	S(x) = - \sum_{c=0}^{255}{ P_{c}(x) * \log{P_{c}(x)} }
% 	\end{equation}
%
% 	then, the metric over a batch of true images $X_{true}$ and a batch of generated images $X_{gen}$, both consisting of $N$ samples, is calculated as:
%
% 	\begin{equation}
% 	Entropy_{mae}(X_{true}, X_{gen}) = \frac{1}{N}\sum_{i=0}^{N-1} | S(X_{gen}(i)) - S(X_{true}(i)) |
% 	\end{equation}

	% This metric is related to how different the entropy of a generated image is from the corresponding real image, which can also be interpreted as the difference in the quantity of information expressed by the two samples. In general, large values of this metric indicates that the generated sample is close to random noise or the topology of the two levels are greatly different, while small values indicates that the entropies of the two images are on a comparable level.

	\medskip\noindent
	\textbf{\similarity{}}:
  we computed the average Structural Similarity (SSIM)~\cite{ssim} index between the images of human-designed and generated levels.
  This metric takes into account several characteristics of the images, such as the luminance, the contrast and the local structure.
  The value \similarity{} is comprised between 0 and 1 (where 1 is achieved when comparing the same images).
  %
	% This metric is defined as the Structural Similarity (SSIM) Index \cite{ssim} between two images.  This measure is the result of a framework that consider several aspects of an image, such as the luminance, the contrast and the structure, rather than basing only on a single statistic. Moreover, the structural similarity technique is applied locally over the image, for reflecting the fact that pixels are more correlated to close pixels than distant ones. \\*
	% For calculating this metric we use the implementation provided by the Scikit-Image Python library, which we leave the formulation to the paper \cite[p.~604]{ssim}, and compute the mean of the SSIM index over the images belonging to the true and generated batches. Regarding the interpretation of the metrics, higher values indicate the fact that true and generated samples are often structurally similar: in other words, the network produces samples in which the local structure of the pixels are comparable.

	\medskip\noindent
	\textbf{\encerr{}}:
  given that in some level images pixel could have only few \emph{meaningful} values (e.g., in the \emph{FloorMap} each pixel should
    have either value 0, black, or 255, white), we computed how far each pixel is to its closest meaningful value.
  Accordingly, we computed the \emph{Encoding Error} of the network as a measure of the average errors over the pixel values of the level images
    generated by the network.

	\medskip\noindent
	\textbf{\cornerr{}}:
  first, we used the Harris detector~\cite{harrisdetector} to compute the number of corners contained in \emph{FloorMap} and \emph{WallMap} images.
  Then, we computed the average \emph{Corner Error}~\cite{slam} between the images of human-designed levels and generated levels, that is a measure of how
    large is the difference between the average number of corners in the two sets of levels.
The resulting metric provides an estimated of the how close are human-designed and generated levels in terms of structural complexity.
 %
 % This error is based on the idea introduced with the corner count metric introduced by \cite{slam}. We define our implementation of the Corner Error of two images as:
	% \begin{equation}
	% C_{err}(n_{x}, n_{y}) = \sqrt{\frac{(n_{x} - n_{y})^2}{n_{x}n_{y}}}
	% \end{equation}
	% where $n_{x}$ and $n_{y}$ are respectively the corner count of two binary images, extracted using the Harris corner detector \cite{harrisdetector}.
	% This formula may seem arbitrary, but it demonstrated to scale well on our dataset, since the corner count is actually limited by the size of our samples. The final metric is computed, as the other cases, averaging the corner error over the images in the batch:

	% \begin{equation}
	% MCE(X_{true}, X_{gen}) = \frac{1}{N} \sum_{i=0}^{N-1} C_{err}(n_{true,i}, n_{gen,i})
	% \end{equation}
	% Again, for simplicity, we indicated as $ n_{true,i} $ and $ n_{gen,i} $ the corner count given by \\ $ n_{i} = count(peak(Harris(X_{i}))) $, with obvious meaning of the function names.
	% This metric is proportional to the average distance that the true and generated samples have, giving a quantitative measure of relative map complexity. It is worth nothing that it's not always the case, since generation artefacts may dramatically increase this value, producing a great number of corners. For this reason, this quantity reflects both the relative average complexity between batches of images and the presence of noise or artefacts in the generated samples.
%%% TEXEXPAND: END FILE ./sec_expdesign.tex

\section{Experimental Results}
\label{sec:results}
%%% TEXEXPAND: INCLUDED FILE MARKER ./sec_results.tex
% !TEX root=main.tex
We trained two GANs, an \emph{unconditional} network using level images and a \emph{conditional} network using also the level features;
	then, we compared them based on the evaluation of the level images they can generate.
\begin{figure}
\begin{center}
\includegraphics[width=\TRAINMETRICWIDTH]{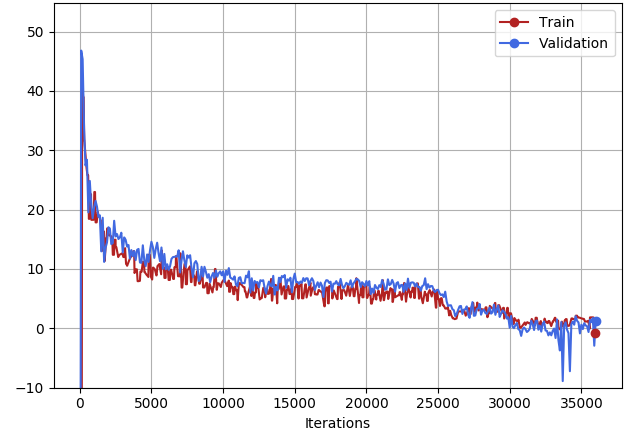}
\end{center}
\caption{The value of discriminator loss of the \emph{unconditional} network during the training process; the red line shows the loss computed
  on the training set, the blue line shows the loss computed on the validation set.}
\label{fig:train-uncond-loss}
\end{figure}
\begin{figure}
\begin{center}
\includegraphics[width=\TRAINMETRICWIDTH]{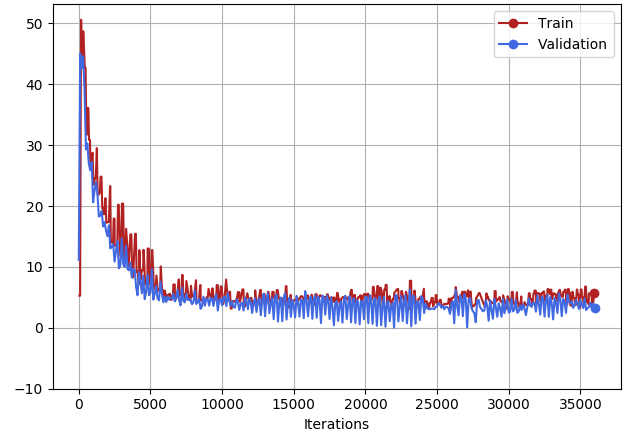}
\end{center}
\caption{The value of discriminator loss of the \emph{conditional} network during the training process; the red line shows the loss computed
  on the training set, the blue line shows the loss computed on the validation set.}
\label{fig:train-cond-loss}
\end{figure}
Figure~\ref{fig:train-uncond-loss} and Figure~\ref{fig:train-cond-loss} show the discriminator loss of the \emph{unconditional} and
  the \emph{conditional} networks.
%%%
The discriminator loss steadily decreases with the training iterations in both types of network;
in addition, the loss achieved on the training set (red line in Figure~\ref{fig:train-uncond-loss} and~\ref{fig:train-cond-loss})
	is close to the one achieved on the validation set (blue line in Figure~\ref{fig:train-uncond-loss} and~\ref{fig:train-cond-loss}), suggesting
	that the networks have good generalization capabilities;
finally, the learning process seems slightly faster for the conditional network than for unconditional one, although, due to the
	adversarial nature of the network, the discriminator loss value alone is not representative of the quality of the generated images.

%
% \begin{figure}
% 	\includegraphics[width=\columnwidth]{entropy.png}
% 	\caption{The value of \emph{entropy} metric computed on the generated images during the training process: the value achieved by the
% 	  \emph{unconditional} network is reported with a red line, the one achieved by the \emph{conditional} network is reported with a blue line;
% 	  the \emph{entropy} value is computed as the average over all the level images (i.e., \emph{FloorMap}, \emph{WallMpa}, \emph{HeightMap},
% 		and \emph{ThingsMap}).}
% 	\label{fig:train-entropy-mean}
% \end{figure}
%
\begin{figure}
	\begin{center}
	\includegraphics[width=\TRAINMETRICWIDTH]{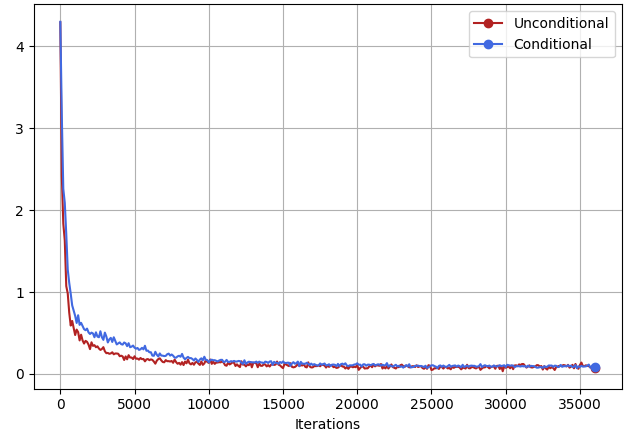}
	\end{center}
	\caption{The value of \entropy{} computed on the \emph{FloorMap} of the levels generated during the training process:
	  \entropy{} achieved by the \emph{unconditional} network is reported with a red line, \entropy{} achieved by the \emph{conditional} network
		is reported with a blue line.}
	\label{fig:train-entropy-floormap}
\end{figure}
\begin{figure}
	\begin{center}
	\includegraphics[width=\TRAINMETRICWIDTH]{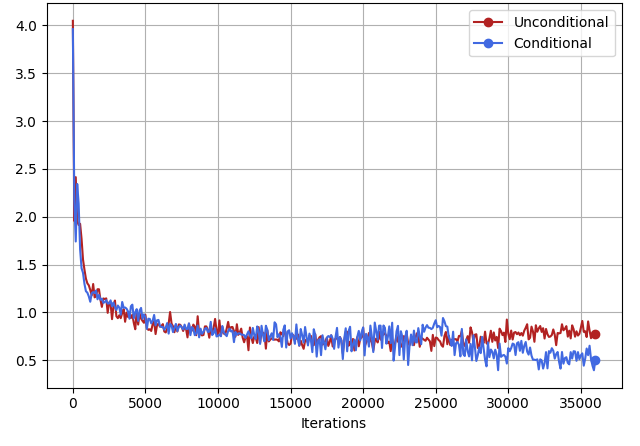}
	\end{center}
	\caption{The value of \entropy{} computed on the \emph{HeightMap} of the levels generated during the training process:
	  \entropy{} achieved by the \emph{unconditional} network is reported with a red line, \entropy{} achieved by the \emph{conditional} network
		is reported with a blue line.}
	\label{fig:train-entropy-heightsmap}
\end{figure}
Figure~\ref{fig:train-entropy-floormap} compares the entropy of the pixel distribution \entropy{} for the
	\emph{FloorMap} images of the levels generated by the \emph{unconditional} and \emph{conditional} networks.
Both networks are able to achieve a \entropy{} value close to 0 at the end of the training;
  	thus, the \emph{FloorMap} of the generated levels have an entropy level very similar to human-designed levels.
The analysis of the \entropy{} value computed on \emph{WallMap} and \emph{ThingsMap} (not reported here) gives similar results, despite
  the final values of \entropy{} are slightly greater as the generated images are slightly more noisy than \emph{FloorMap}.
%%%
Figure~\ref{fig:train-entropy-heightsmap} shows the \entropy{} computed on the \emph{HeightMap} of the generated levels:
although the \emph{conditional} network is able to achieve slightly smaller \entropy{} (at the end of the training)
  than \emph{unconditional} network, none of the two networks is able to achieve a \entropy{} value close to 0;
the result is not surprising as in human-designed levels there is usually a very limited number of height levels, while the generated
  \emph{HeightMap} images are much more noisy.

\begin{figure}
	\begin{center}
	\includegraphics[width=\TRAINMETRICWIDTH]{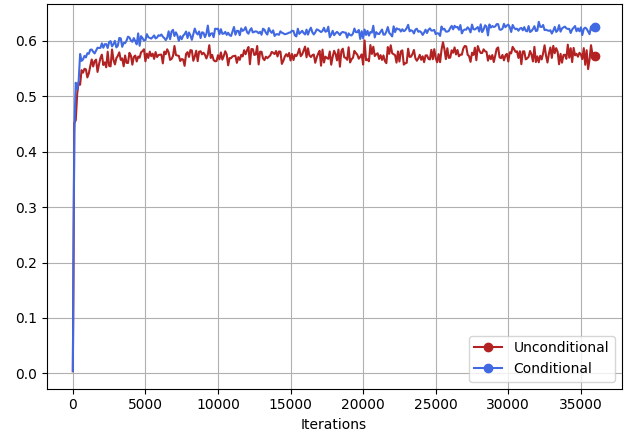}
	\end{center}
	\caption{The value of \similarity{} computed on the levels generated during the training process: \similarity{} achieved by the \emph{unconditional}
	  network is reported with a red line, \similarity{} achieved by the \emph{conditional} network is reported with a blue line;
		the curves are computed as the average of the \similarity{} computed on  \emph{FloorMap}, \emph{WallMap}, \emph{HeightMap},	and \emph{ThingsMap}.}
	\label{fig:train-similarity}
\end{figure}
Figure~\ref{fig:train-similarity} compares the average Structural Similarity index \similarity{} computed on the levels generated by the two networks;
the results show that both networks  rapidly improve the \similarity{} of the generated levels at the beginning of the training process;
in addition, they also suggest that the levels generated by \emph{conditional} network have an overall quality that is slightly better than
	the one generated by the \emph{unconditional} network.

\begin{figure}
	\begin{center}
	\includegraphics[width=\TRAINMETRICWIDTH]{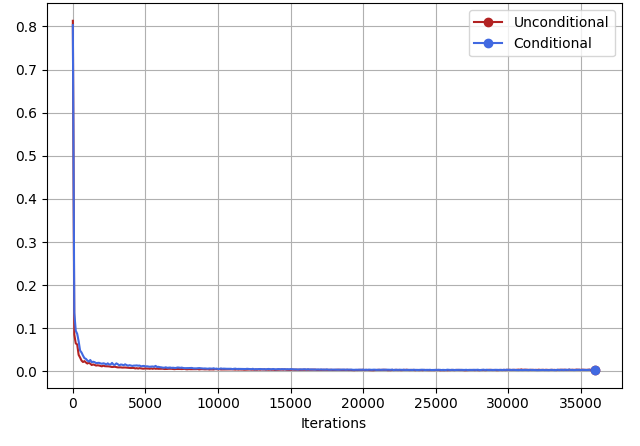}
	\end{center}
	\caption{The value of \encerr{} computed on the \emph{FloorMap} of the levels generated during the training process: \encerr{} achieved by
	    the \emph{unconditional} network is reported with a red line, \encerr{} achieved by the \emph{conditional} network is reported with a blue line.}
		\label{fig:train-encoding_error-floormap}
	\end{figure}
The analysis of the network Encoding Error \encerr{} computed on the level images generated by the two networks shows that both networks
	quickly reduce EE during the training process achieving almost the same final performance at the end.
%%%
Figure~\ref{fig:train-encoding_error-floormap} shows the \encerr{} computed for the \emph{FloorMap} of the generated levels;
both networks generate images with very small errors at the end of the training process.
%such a result allows to apply a very simple post-processing to automatically fix \emph{wrong} pixels in the
%  generated images, i.e., the pixel are assigned to the closest value among 0 and 255. This post-processing is very effective for \emph{FloorMap} images.

\begin{figure}
	\begin{center}
	\includegraphics[width=\TRAINMETRICWIDTH]{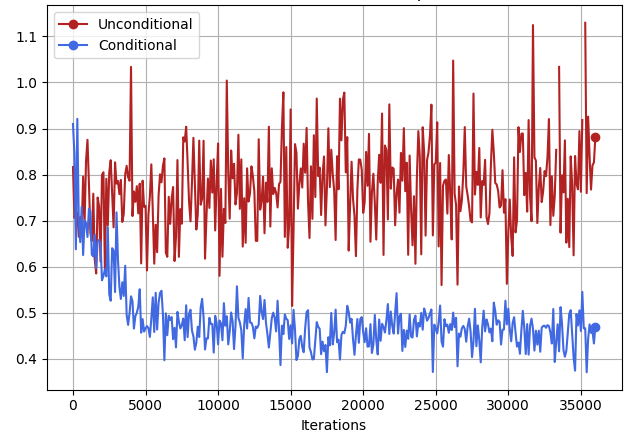}
	\end{center}
	\caption{The value of \cornerr{} computed on the \emph{FloorMap} of the levels generated during the training process: \cornerr{} achieved by
	    the \emph{unconditional} network is reported with a red line, \cornerr{} achieved by the \emph{conditional} network is reported with a blue line.}
		\label{fig:floor_corner_error}
\end{figure}
\begin{figure}
	\begin{center}
	\includegraphics[width=\TRAINMETRICWIDTH]{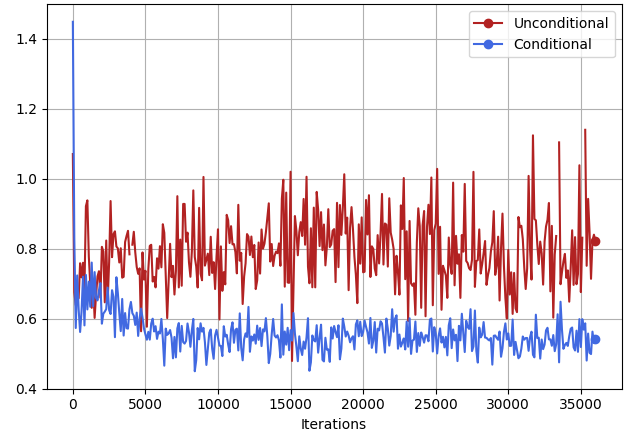}
	\end{center}
	\caption{The value of \cornerr{} computed on the \emph{WallMap} of the levels generated during the training process: \cornerr{} achieved by
	    the \emph{unconditional} network is reported with a red line, \cornerr{} achieved by the \emph{conditional} network is reported with a blue line.}
		\label{fig:wall_corner_error}
\end{figure}
Figure~\ref{fig:floor_corner_error} and Figure~\ref{fig:wall_corner_error} show the average Corner Error \cornerr{} computed respectively on the \emph{FloorMap} and on the \emph{WallMap} of the levels generated by the two networks;
the results suggest that the \cornerr{} of the levels generated by the \emph{unconditional} network does not improve with training, while
  it does (at least slightly) in the levels generated by \emph{conditional} network;
in addition, during the whole training process, the \cornerr{} appears to be much less noisy for the \emph{unconditional} network;
these results suggest that, overall, levels generated by \emph{conditional} network have a structural complexity more similar to the
	human-designed levels.

	\begin{figure}
	\begin{minipage}{\columnwidth}
		\begin{center}
			\includegraphics[width=2cm]{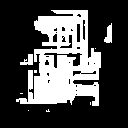}
			\includegraphics[width=2cm]{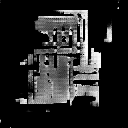}
			\includegraphics[width=2cm]{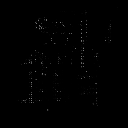}
			\includegraphics[width=2cm]{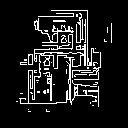}
		\end{center}

		\begin{center}
			\includegraphics[width=2cm]{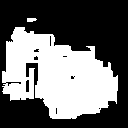}
			\includegraphics[width=2cm]{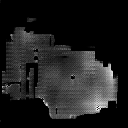}
			\includegraphics[width=2cm]{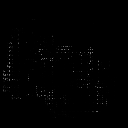}
			\includegraphics[width=2cm]{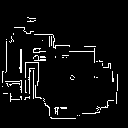}
		\end{center}

		\begin{center}
			\includegraphics[width=2cm]{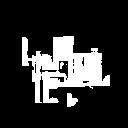}
			\includegraphics[width=2cm]{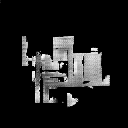}
			\includegraphics[width=2cm]{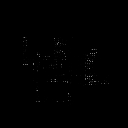}
			\includegraphics[width=2cm]{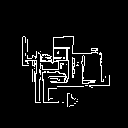}
		\end{center}

	\end{minipage}
	\caption{Samples generated by the unconditional network. From left to right: the \emph{FloorMap}, \emph{HeightMap}, \emph{ThingsMap}, and \emph{WallMap} of the generated levels.}
	\label{fig:samples-uncond}
	\end{figure}
	\begin{figure}
		\begin{minipage}{\columnwidth}

	    \begin{center}
			    \includegraphics[width=2cm]{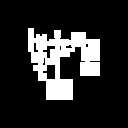}
	        \includegraphics[width=2cm]{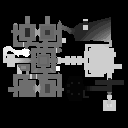}
	        \includegraphics[width=2cm]{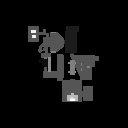}
	    \end{center}

	    \begin{center}
	      \includegraphics[width=2cm]{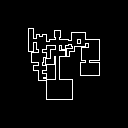}
	      \includegraphics[width=2cm]{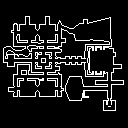}
	      \includegraphics[width=2cm]{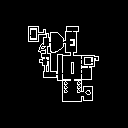}
	    \end{center}

	    \begin{center}
	      \includegraphics[width=2cm]{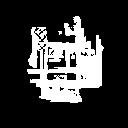}
	      \includegraphics[width=2cm]{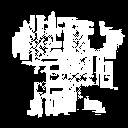}
	      \includegraphics[width=2cm]{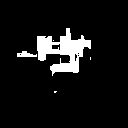}
	    \end{center}

			\begin{center}
	  		\includegraphics[width=2cm]{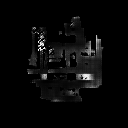}
	      \includegraphics[width=2cm]{sample13_map_heightmap_generated.png}
	      \includegraphics[width=2cm]{sample13_map_heightmap_generated.png}
	    \end{center}

	    \begin{center}
	  		\includegraphics[width=2cm]{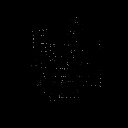}
	      \includegraphics[width=2cm]{sample13_map_thingsmap_generated.png}
	      \includegraphics[width=2cm]{sample13_map_thingsmap_generated.png}
	    \end{center}

	    \begin{center}
	  		\includegraphics[width=2cm]{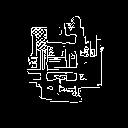}
	      \includegraphics[width=2cm]{sample13_map_wallmap_generated.png}
	      \includegraphics[width=2cm]{sample13_map_wallmap_generated.png}
		   \end{center}

		\end{minipage}
		\caption{Samples generated by the conditional network.
	    In each column, from the top to the bottom, are reported (i) the \emph{HeightMap} and (ii) the \emph{WallMap} of the DOOM
			  level used to extract the features in input to the network, (iii) the \emph{FloorMap}, (iv) the \emph{HeightMap}, (v) the
				\emph{ThingsMap}, and (vi) the \emph{WallMap} of the level generated by the network.}
		\label{fig:samples-cond}
	\end{figure}

Finally, Figure~\ref{fig:samples-uncond} and Figure~\ref{fig:samples-cond} report examples of levels generated by the two networks.
The visual comparison shows that the levels generated by the \emph{conditional} network (Figure~\ref{fig:samples-cond})
 	have a richer structure and, overall, look slightly more similar to the human-designed levels.
On the other hand, the images suggest that the networks struggle  to reproduce smaller details of the levels, probably
	due to generation noise.
%%% TEXEXPAND: END FILE ./sec_results.tex

\section{Conclusions}
\label{sec:conclusions}
%%% TEXEXPAND: INCLUDED FILE MARKER ./sec_conclusions.tex
We trained two Generative Adversarial Networks (GANs)
	to learn a model from more than 1000 DOOM levels.
The unconditional model was trained with level images extracted from the original WAD files.
The conditional model was trained using the same images and additional topological features extracted from the level structure.
We applied the models to generate new levels and analyzed the types of levels generated with and without the additional features.
%%%
Our results show that the two networks can generate new levels with characteristics that are
	similar to the levels used for training.
They also show that the additional features used to train the conditional network increase the quality of generated samples and lead to better learning.
%, while also providing a method to influence the network output during the sampling process.
%%%
%One of the most common concerns when working with GANs is that the network could overfit the training set, learning to reproduce samples from the dataset. While we cannot prove it formally, we refer to the results in \cite{empiricalevaluation}, which claim that overfit is difficult to occur in the type of model we used, even for a small number of training samples, de facto demonstrating that the behaviour of GANs is different from that of classical deep neural networks used for classification.
%%%
The evaluation of samples generated with GANs is still a recent field of research and
so far there is no prevailing approach. Moreover, our domain makes is difficult to apply the commonly used methods to assess sample quality. In this work, we proposed a qualitative method for assessing the generated sample quality during the training process that works for DOOM level images.
%The method we proposed succeeds in indicating that the network is actually learning the level structures, the metrics we proposed have the drawback that they need to be calculated on each map differently in order to benefit of their informational power, while considered altogether for assessing the general sample quality.

Our promising results, although preliminary, represent an excellent
	starting point for future improvements and highlight
	a viable alternative to classical procedural generation.
%%%
Most generated levels have proved to be interesting to explore and play due to the presence of typical features of DOOM maps
	(like narrow tunnels and large rooms).
%%%
Our approach extracts domain knowledge through learning and so it does not require an expert to encode it explicitly,
	as traditional procedural generation often does.
Thus, human designers can focus on high-level features by including specific types of maps or features
	in the training set as those we selected as network inputs.
\end{document}